\documentclass[review,12pt]{elsarticle}

\usepackage[margin=1in]{geometry} 
\usepackage{setspace}             
\usepackage{parskip}              
\usepackage{amssymb}
\usepackage{amsmath}
\usepackage{bm}
\usepackage{booktabs}
\usepackage{multirow}
\usepackage{hyperref}

\usepackage{graphicx}
\usepackage{subcaption}

\journal{Artificial Intelligence in Medicine}

\newcommand{\fixed}{f}
\newcommand{\moving}{m}
\newcommand{\deform}{\boldsymbol{\phi}}
\newcommand{\disp}{\boldsymbol{u}}

\newcommand{\dnet}{\boldsymbol{\Phi}}
\newcommand{\unet}{\boldsymbol{U}}

\newcommand{\sun}{\mathcal{E}_{\textnormal{sim}}}
\newcommand{\run}{\mathcal{E}}
\newcommand{\gun}{\mathcal{R}}
\newcommand{\energy}{\mathrm{E}}

\newcommand{\strain}{\eta}
\newcommand{\trace}{\operatorname{trace}}
\newcommand{\wadapt}{\hat \alpha}
\newcommand{\wvol}{\hat \lambda}
\newcommand{\wshear}{\hat \mu}

\newcommand{\sigmoid}{\sigma}

\newcommand{\norm}[1]{\left\lVert#1\right\rVert}

\usepackage{enumitem}
\usepackage{pifont} 
\usepackage{xcolor} 

\newenvironment{myitem}
  {\begin{itemize}[label=\textcolor{gray!70}{\ding{110}}, leftmargin=1.5em, labelsep=0.6em,
    itemsep=0.5em, topsep=0.3em, parsep=0pt, partopsep=0pt]}
  {\end{itemize}}

\allowdisplaybreaks
  
\begin{document}

\begin{frontmatter}

\title{DARE: A Deformable Adaptive Regularization Estimator for Learning-Based Medical Image Registration}

 \author[label1]{Ahsan Raza Siyal}
 \author[label1]{Markus Haltmeier}
 \author[label2,label3]{Ruth Steiger}
 \author[label2,label3]{Malik Galijasevic}
 \author[label2,label3]{Elke Ruth Gizewski}
 \author[label2,label3]{Astrid Ellen Grams}

 \affiliation[label1]{organization={Department of Mathematics, University of Innsbruck},
             country={Austria}}

 \affiliation[label2]{organization={Department of Radiology, Medical University of Innsbruck},
             country={Austria}}

 \affiliation[label3]{organization={Neuroimaging Research Core Facility, Medical University of Innsbruck},
             country={Austria}}

\begin{abstract}
Deformable medical image registration is a fundamental task in medical image analysis. While deep learning-based methods have demonstrated superior accuracy and computational efficiency compared to traditional techniques, they often overlook the critical role of regularization in ensuring robustness and anatomical plausibility. We propose DARE (Deformable Adaptive Regularization Estimator), a novel registration framework that dynamically adjusts elastic regularization based on the gradient norm of the deformation field. Our approach integrates strain and shear energy terms, which are adaptively modulated to balance stability and flexibility. To ensure physically realistic transformations, DARE includes a folding-prevention mechanism that penalizes regions with negative deformation Jacobian. This strategy mitigates non-physical artifacts such as folding, avoids over-smoothing, and improves both registration accuracy and anatomical plausibility.
\end{abstract}

\begin{keyword}
Medical image registration  \sep  
deep learning  \sep 
adaptive regularization \sep
physics informed network \sep
 \end{keyword}

\end{frontmatter}
 
\section{Introduction}

Deformable image registration plays a pivotal role in medical image analysis, serving as a cornerstone for numerous clinical applications. By aligning a moving image with a fixed image through non-linear transformations, deformable registration establishes a dense correspondence between the two, enabling precise spatial alignment of anatomical or pathological features. This technique facilitates critical tasks for healthcare professionals, including medical imaging \cite{zhou2024deep}, brain mapping \cite{wang2021bi, siyal}, identifying and localizing lesions \cite{li2014accurate}, treatment planning and procedural guidance \cite{lu2010three}, evaluating the effectiveness of therapies \cite{holden2002quantification}, and monitoring disease activity and progression \cite{meier2003time}.

\subsection{Classical and learned image registration}

Classical deformable image registration \cite{Avants2008, Vercauteren2009, rueckert1999nonrigid, scherzer2006mathematical, hinterberger2001models, Heinrich2013, droske2004variational, poschl2010variational, sotiras2013deformable, weickert2001variational, modersitzki2003numerical} predominantly relies on variational and iterative optimization techniques that balance the similarity between moving and fixed images with the regularity of the deformation field. To mitigate overfitting and promote anatomically plausible transformations, several methods have been developed to constrain the registration function to remain close to a diffeomorphism \cite{bajcsy1989multiresolution, rueckert1999nonrigid, vialard2012diffeomorphic, schmah2013left, yang2017quicksilver, wang2020deepflash, shen2019networks}. These include Large Deformation Diffeomorphic Metric Mapping (LDDMM) \cite{Beg2005}, vector Stationary Velocity Field (vSVF) models \cite{shen2019networks}, B-spline-based registration \cite{rueckert1999nonrigid}, elastic registration \cite{bajcsy1989multiresolution}, and Demons-based techniques \cite{Vercauteren2009}. While effective in preserving topological consistency and enabling smooth deformations, these approaches often involve computationally expensive and iterative numerical optimization processes, limiting their practical applicability in time-sensitive clinical workflows.

In recent years, the emergence of learning-based  methods~\cite{Balakrishnan2018, Balakrishnan2019, Jia2022, tony, Qiu2022, Dalca2019, Kim2021, Lei2020, Chen2022, zhou2023self, hemon2024indirect} has revolutionized deformable image registration. These approaches aim to learn an explicit registration function, thereby eliminating the need for repetitive optimization during inference. They significantly reduce computational overhead while achieving performance that often rivals traditional techniques. However, deep learning-based registration methods typically prioritize maximizing similarity metrics between the transformed moving image and the fixed image. This focus can lead to overfitting to the similarity measure, resulting in non-smooth or anatomically implausible deformations---an undesirable outcome in medical imaging, where smooth and coherent transformations are essential. To address this, recent studies have incorporated regularization terms such as diffusion regularization~\cite{Balakrishnan2018, Balakrishnan2019, hoffmann2021synthmorph, Jia2022, Chen2022, wang2023modet, chang2024cgnet} and bending energy regularization~\cite{rueckert1999nonrigid, johnson2001landmark, Chen2022, wolterink2022implicit, wu2020align} to constrain the deformation field. These regularization terms are critical for ensuring smooth and plausible transformations. However, they rely on uniform regularization parameters, which can lead to over-smoothing in regions requiring detailed alignment or flexibility, and they fail to capture the local variations critical for high-precision tasks in medical imaging.

\subsection{Proposed Deformable Adaptive Regularization}

To address  limitations of existing methods, we propose a Deformable Adaptive Regularization Estimator (DARE)  for learned  deformable medical image registration. DARE follows the  unsupervised learning paradigm and integrates dynamic adjustment of elastic regularization parameters based on the gradient of the deformation field, allowing the regularization strength to adapt spatially and contextually to the characteristics of the deformation.  DARE incorporates strain and shear energy terms, which are adaptively modulated to ensure a balance between stability and flexibility in the deformation field. This adaptive modulation allows the model to handle complex deformations while preserving structural detail, especially in heterogeneous regions where sharp transitions or localized changes are required.  Additionally, DARE integrates a folding prevention mechanism, which penalizes regions where the deformation Jacobian determinant becomes negative. This ensures that the deformation field remains invertible and physically realistic, addressing a critical shortcoming of many traditional approaches that may allow non-physical folding artifacts.  

DARE minimizes over-smoothing in regions that demand flexibility while maintaining smoothness and anatomical plausibility in less variable areas. This also prevents the loss of critical anatomical details, such as small lesions or fine structures, which are often obscured under uniform regularization schemes. Furthermore, the dynamic nature of the proposed method ensures that the regularization strength is context-sensitive, allowing the registration framework to accommodate the diverse and complex deformation patterns commonly encountered in medical imaging. DARE significantly enhances the robustness and dependability of learning-based registration models, ensuring consistent and precise performance across diverse datasets. By addressing the identified shortcomings of traditional diffusion and bending energy regularizations, DARE delivers improved accuracy, anatomical plausibility, and physical realism, making it particularly well-suited for applications in medical contexts requiring high precision and consistency.

\medskip
Our  primary contributions are as follows:

\begin{myitem}

    \item \textbf{Adaptation to Deformation Complexity:} 
    DARE integrates adaptive strain and shear regularization terms whose strength is derived from the gradient norm of the displacement field. By monitoring local gradient magnitudes, DARE stiffens the transformation in regions with abrupt displacement changes—preserving anatomical edges—while relaxing the regularization in smoother areas, enabling the model to bend and twist where the data demand it.

    \item \textbf{Volume Change Regulation in Key Structures:} 
    Through dynamic adjustment of regularization parameters, DARE mitigates unrealistic volume changes in critical anatomical regions. This is supported by quantitative experiments demonstrating reduced volume distortions, thereby enhancing anatomical plausibility.

    \item \textbf{Adaptive Regularization Weighting:} 
    We propose a deformation-aware weighting scheme that automatically modulates the smoothness penalty based on local displacement complexity. In gently varying areas, the constraint is relaxed, allowing the deformation to better conform to fine anatomical structures.

    \item \textbf{Favorable Accuracy--Plausibility Trade-off:} 
    Across multiple 3D MRI benchmarks, DARE consistently reduces registration error while substantially decreasing the number of folding voxels. This demonstrates that its physics-informed design achieves state-of-the-art alignment accuracy without compromising deformation realism.
\end{myitem}

\subsection{Related Work}

Unsupervised registration methods, which eliminate the need for ground truth deformation fields, have garnered significant attention in the field of image registration. Balakrishnan et al.~\cite{Balakrishnan2018} introduced VoxelMorph, a deformable registration method that leverages convolutional neural networks (CNNs) using a U-Net-like~\cite{Ronneberger2015}
Transformers, based on the attention mechanism, were initially introduced for machine translation tasks \cite{vaswani2017attention}. Their ability to reduce the inductive biases of convolutions and capture long-range dependencies has made them highly relevant in computer vision \cite{Dosovitskiy2020, Liu2021}, where they excel in feature extraction \cite{10595015, wang2023modet}. A dual transformer network (DTN) \cite{Zhang2021}, built on a 5-layer U-Net for diffeomorphic image registration, significantly increases computational costs and GPU memory usage. Subsequent methods \cite{Jia2022, zhao} adopted cascaded U-Nets for a coarse-to-fine approach, improving performance but limiting feasibility on low-end GPUs. Later advancements, such as ViT-V-Net \cite{Chen2021} and TransMorph \cite{Chen2022}, integrated Vision Transformer (ViT) and Swin-Transformer blocks to enhance computational efficiency and registration accuracy. Despite these innovations, we show that U-Net variations remain competitive in medical image registration.

Initial learning-based registration methods often focus on maximizing similarity metrics between transformed moving and fixed images. While effective, this approach can lead to overfitting and non-smooth transformations—an issue particularly critical in medical imaging, where smooth and anatomically coherent mappings are essential.  To address this, various techniques have been developed to enforce diffeomorphism. For example, LDDMM~\cite{Beg2005} ensures smooth transformations but is computationally expensive; vSVF~\cite{shen2019networks} struggles with complex deformations; B-spline methods~\cite{rueckert1999nonrigid} require extensive parameter tuning; and Elastic and Demons-based techniques~\cite{bajcsy1989multiresolution, Vercauteren2009} balance smoothness and accuracy but may falter in handling complex scenarios. 

Recent learning-based approaches incorporate diffusion and bending energy  to constrain the deformation fields. Diffusion regularization~\cite{Balakrishnan2018, Balakrishnan2019, hoffmann2021synthmorph, Jia2022, Chen2022, wang2023modet, chang2024cgnet} aims to penalize the gradient of the displacement field, promoting smooth and gradual changes across the image domain. This ensures that deformations do not exhibit abrupt shifts or unnatural distortions. However, a significant disadvantage of diffusion regularization is its tendency to over-smooth deformations, which can suppress sharp transitions at anatomical boundaries. This may lead to inaccuracies when precise alignment of structures with high contrast, such as bones or organ edges, is required. Furthermore, diffusion regularization lacks physical interpretability because it does not account for the biomechanical properties of tissues, making the resulting deformations less realistic in medical applications. It also struggles to handle complex, localized deformation patterns in high-resolution datasets and does not inherently guarantee invertibility of the deformation field, which is critical in many scenarios such as inverse consistency in medical imaging.

Bending energy registration~\cite{rueckert1999nonrigid, johnson2001landmark, Chen2022, wolterink2022implicit, wu2020align}, on the other hand, penalizes the second-order derivatives of the displacement field to ensure the deformation field is not only smooth but also gently bending. It reduces sharp curvature in the deformation field, making it more uniform and avoiding abrupt transitions. However, the bending energy term can impose excessive stiffness on the deformation field, limiting its ability to adapt to complex anatomical changes. In regions where sharp transitions or highly localized changes are required, such as in areas with small or irregular anatomical features, this method can overly constrain the deformation, resulting in inaccuracies. Additionally, the computational cost of calculating and optimizing second-order derivatives is significantly higher compared to diffusion regularization, particularly for high-dimensional medical imaging data. In the context of deep learning frameworks, the use of bending energy regularization introduces numerical complexity, which can lead to instability during model optimization and training.

The stationary nature makes it difficult to capture local variations, resulting in suboptimal transformations that underestimate anatomical intricacies. The can result in excessive smoothing in areas where precise alignment or greater flexibility is needed, thereby failing to capture essential local variations crucial for high-accuracy medical imaging. To overcome these challenges, we propose a dynamic adjustment mechanism for elastic regularization, where the strength of regularization adapts spatially and contextually based on the gradient norm of the deformation field, ensuring more accurate and context-aware alignments.

\subsection{Outline}

The rest of the paper is organized as follows.  In Section~\ref{sec:methods}, we introduce we formulate the deformable registration problem  and describe the proposed learned adaptive image registration method using DARE. Results and comparisons with existing methods are presented in Section~\ref{sec:results}. The paper ends with a short discussion and summary presented in Section~\ref{sec:conclusion}.

\section{Methods}
\label{sec:methods}

Deformable Adaptive Regularization Estimator (DARE), a physics-informed adaptive regularization method, is a critical technique for deformable image registration because it promotes realistic, stable deformations while preserving flexibility in the deformation field. This approach combines elastic energy regularization, gradient-based adaptive weighting, and folding prevention into a unified framework.

\subsection{Deformable Image Registration}

Let \(\moving, \fixed \colon \Omega \to \mathbb{R}\) denote two given 3D images (volumes), where the first (the moving image) is to be deformed to match the second (the fixed image). The goal of deformable image registration is to compute a deformation field \(\deform \colon \Omega \to \Omega\) such that the transformed moving image \(\moving \circ \deform\) aligns closely with the fixed image \(\fixed\), while ensuring that the deformation field remains regular and physically plausible.

Throughout, we write \(\deform(x) = x + \disp(x)\) for \(x \in \Omega\), where \(\disp(x)\) is the displacement field.
In classical deformable image registration, the displacement field \(\disp\) is computed independently for each image pair \((\fixed, \moving)\) by minimizing an objective function that balances image similarity with deformation regularity. In contrast, learning-based approaches aim to construct a parameterized function that maps pairs of fixed and moving images to the corresponding displacement field. Once trained, such a model eliminates the need for time-consuming iterative optimization and allows for the exploitation of shared structural patterns across different image pairs.

Suppose the parametric function predicting the displacement field is drawn from a network architecture \(\unet_\theta \), where \(\theta \in \Theta\) denotes the learnable parameters. When ground truth displacement fields are available, the network can be trained in a supervised manner.  However, in most real-world applications, such ground truth is unavailable, and only pairs \((\fixed_i, \moving_i)\) of fixed and moving images are provided. In this unsupervised learning setting, the deformation \(\dnet_\theta = \mathrm{Id} + \unet_\theta\) is trained by minimizing the unsupervised registration functional
\begin{equation} \label{eq:unsupervised}
\run( \theta ) 
\triangleq
\sum_{i} \sun \big(\fixed_i, \moving_i \circ \dnet_\theta(\fixed_i, \moving_i)\big)
+ \alpha\, \gun\big(\unet_\theta(\fixed_i, \moving_i)\big) \,,
\end{equation}
where the similarity term \(\sun\) quantifies the alignment between the fixed image $\fixed_i$ and the warped moving image
$\moving_i \circ \dnet_\theta(\fixed_i, \moving_i)$, the regularization term \(\gun\) promotes regularity and discourages anatomically implausible deformations, and the regularization parameter \(\alpha\) that balances image similarity against deformation regularity.

This unsupervised approach provides a robust framework for deformable image registration in scenarios where labeled data is scarce or unavailable. The performance of \eqref{eq:unsupervised} is determined by the quality of the training pairs \((\fixed_i, \moving_i)\), the choice of the similarity and regularization terms, the network architecture, and the optimization algorithm used for its minimization. Optimization is typically performed using stochastic gradient descent or one of its variants, as we follow here.

\subsection{Elastic Energy Regularizer}

Elastic energy regularization ensures that the deformation field remains smooth and physically plausible.  The elastic energy also referred to as strain energy represents the energy stored in a deformation field resulting from the spatial transformations applied to align two images.  It quantifies how much the deformation stretches or compresses different regions of the image during the registration process.

The strain tensor measures the relative displacement of particles in the material and is defined as: is defined by  
\(	 \strain    
	 \triangleq
	 ( \nabla \disp  + \nabla \disp ^T ) /  2
\)
and used in continuum mechanics to describe the deformation of a material. Then, for small deformations, the elastic energy  (or strain  energy) regularizer is given by  
\begin{equation} \label{eq:strain}
	\gun_{\textnormal{elastic}} (\disp) 
	=
	 \int_{\Omega} \left( \lambda_{\textnormal{strain}}   \cdot    \trace( \strain (x))^2 
	+   \mu_{\textnormal{shear}}  \cdot   \norm{  \strain (x) }_F^2 \right) d x  
	\quad \text{ with } \strain    
	 =
	 ( \nabla \disp  + \nabla \disp ^T ) /  2 \,.
\end{equation} 
Here  $\trace ( \strain )  \triangleq \sum_{i=1}^3 \strain_{i,i}$   and $\norm{ \strain }_F   \triangleq (\sum_{i,j=1}^3 \strain_{i,j}^2)^{1/2}$  are the trace   and the  Frobenius norm of the strain tensor, respectively, and   $\lambda_{\textnormal{strain}}$ and $\mu_{\textnormal{shear}}$ are the Lamé parameters. 
Functional $\gun_{\textnormal{elastic}}$ represents the  stored energy resulting from the   spatial deformation field applied to align two images. 
It depends  how much the deformation stretches or compresses different regions of the image during the registration process.  

Each term in \eqref{eq:strain} has a clear physical interpretation. Note that we adopt the term strain energy to refer to the volumetric (dilatational) component of the elastic energy, and shear energy to denote the deviatoric (shape-changing) component.  The strain energy density, \( \energy_ {\textnormal{strain}} \triangleq \lambda_{\textnormal{strain}} \cdot \trace(\strain)^2 \), represents the energy associated with pure volumetric changes—either expansion or compression. It penalizes deviations from volume preservation, with the Lamé parameter \( \lambda_{\textnormal{strain}} \) governing the material’s resistance to volume change (i.e., its compressibility). This term increases when a material element changes in size but not shape. 

The shear energy density, \( \energy_ {\textnormal{shear}} \triangleq \mu_{\textnormal{shear}} \cdot \norm{\strain}_F^2 \), captures the energy due to shape-changing deformations such as shear and distortion induced by the deformation field. The shear modulus \( \mu_{\textnormal{shear}} \) controls the material’s resistance to such distortions. Unlike strain energy, which is tied to volumetric changes, shear energy reflects shape changes at constant volume.

\subsection{Deformable Adaptive Regularizer}

Classically the Lamé  parameters $\lambda_{\textnormal{strain}}$ and $\mu_{\textnormal{shear}}$ are constant, whereas  in our model, we allow the Lamé parameters \( \lambda_{\textnormal{strain}} \) and \( \mu_{\textnormal{shear}} \) to vary spatially and tae them as   functions of the deformation gradient. This extension is particularly well-suited for registration  biological tissue, where mechanical properties such as stiffness and compressibility are highly heterogeneous and dependent on location and local deformation. For example, tumors, scar tissue, or fibrotic regions often exhibit increased stiffness compared to surrounding healthy tissue. By making \( \lambda_{\textnormal{strain}} \) and \( \mu_{\textnormal{shear}} \) spatially adaptive, the model can better capture such variations and provide more physiologically realistic deformation fields, which is especially valuable in image registration and simulation of soft tissue mechanics.  To define he overall regularization we also allow the regularization parameter to be spatially adaptive.

For the proposed  spatial adaptive energy all parameters are dynamically adjusted based on the gradient norm we arrive at 
\begin{equation} \label{eq:dare-reg}
	\gun_{\textnormal{DARE}}(\disp)  
	\triangleq
	  \int_{\Omega} \wadapt(\disp) \cdot \left( \wvol(\disp)   \cdot   \trace(\strain (\disp) )^2 
	+   \wshear (\disp)   \cdot  \norm{  \strain(\disp)}_F^2 \right)  d x \,,
\end{equation}
where 
\begin{align} \label{eq:w-vol}
\wvol(\disp) &\triangleq \lambda_0 \cdot \left( 1 + \Delta \cdot \exp\bigl(-\norm{\nabla \disp} / \theta \bigr) \right)
\\ \label{eq:w-shear}
\wshear (\disp) &\triangleq \mu_0 \cdot  \left( 1 + \Delta \cdot \sigmoid \bigl( -( \norm{\nabla \disp} - \tau)/\kappa \bigr) \right)
\\
\label{eq:w-adapt}
\wadapt (\disp) &\triangleq  1 + \beta_0  \cdot  \exp\bigl( -\norm{\nabla \disp} \bigr ) \,.
\end{align}
Here  $\lambda_0, \mu_0$ are base values; $\theta$  determines sensitivity to deformation gradients;   $\Delta$ controls the gradient-based adjustments;  and $\tau$, $\kappa$ define the shear modulus adaptation  with sigmoid function  $\sigmoid (x) = 1/(1 + \exp(-x))$.

All adaptive parameters are selected based on the deformation gradient norm $\norm{\nabla \disp}$. Specifically they have to following interpretations.   

\begin{myitem}
\item \textbf{Adaptive first Lamé parameter:}
If the magnitude of deformation $\norm{\nabla \disp}$ is small, the exponential term in \eqref{eq:w-vol} is large, leading to larger  $\lambda_{\textnormal{strain}} = \wvol (\disp)$. Conversely if the deformation is large, the exponential term becomes small, resulting in a smaller $\lambda_{\textnormal{strain}}$. The parameter $\Delta$ weights the exponential term, controlling how strongly $\lambda_{\textnormal{strain}}$ changes  to variations in $\norm{\nabla \disp}$. The parameter $\theta$  normalizes the gradient norm and determines the shape and with of $\wvol$. Larger  values make the adjustment smoother, while smaller values make it more sensitive to small changes in the deformation.

\item \textbf{Adaptive shear modulus:}
The adaptive adjustment of $\mu_{\textnormal{shear}} = \wshear(\disp)$ allows the shear modulus to vary smoothly based on the deformation gradient norm $\norm{\nabla \disp}$, enabling finer control of regularization. The parameter $\tau$ shifts the sigmoid's midpoint, determining the gradient norm value where the adjustment is most sensitive. This balance ensures that stronger penalties are applied in regions with significant deformation (above the center), while preserving flexibility in low-deformation regions.

\item \textbf{Adaptive regularization parameter:}  
Dynamically adjusting the regularization parameter $\alpha = \wadapt(\disp)$ provides localized flexibility and enhanced stability. In regions with low gradient norms (smooth areas), the regularization parameter decreases, allowing more flexibility in the deformation field. Conversely, in high-gradient regions (sharp transitions or distortions), the regularization parameter increases allowing to maintain smoothness and physical plausibility. The parameter $\beta$ scales the sensitivity of the weighting factor, amplifying the contrast between weights in high-gradient and low-gradient regions. This ensures efficient regularization, balancing accuracy, and stability by adapting to the local characteristics of the deformation field. 
\end{myitem}

\begin{figure}[htbp]
\centering
\includegraphics[width=0.8\linewidth]{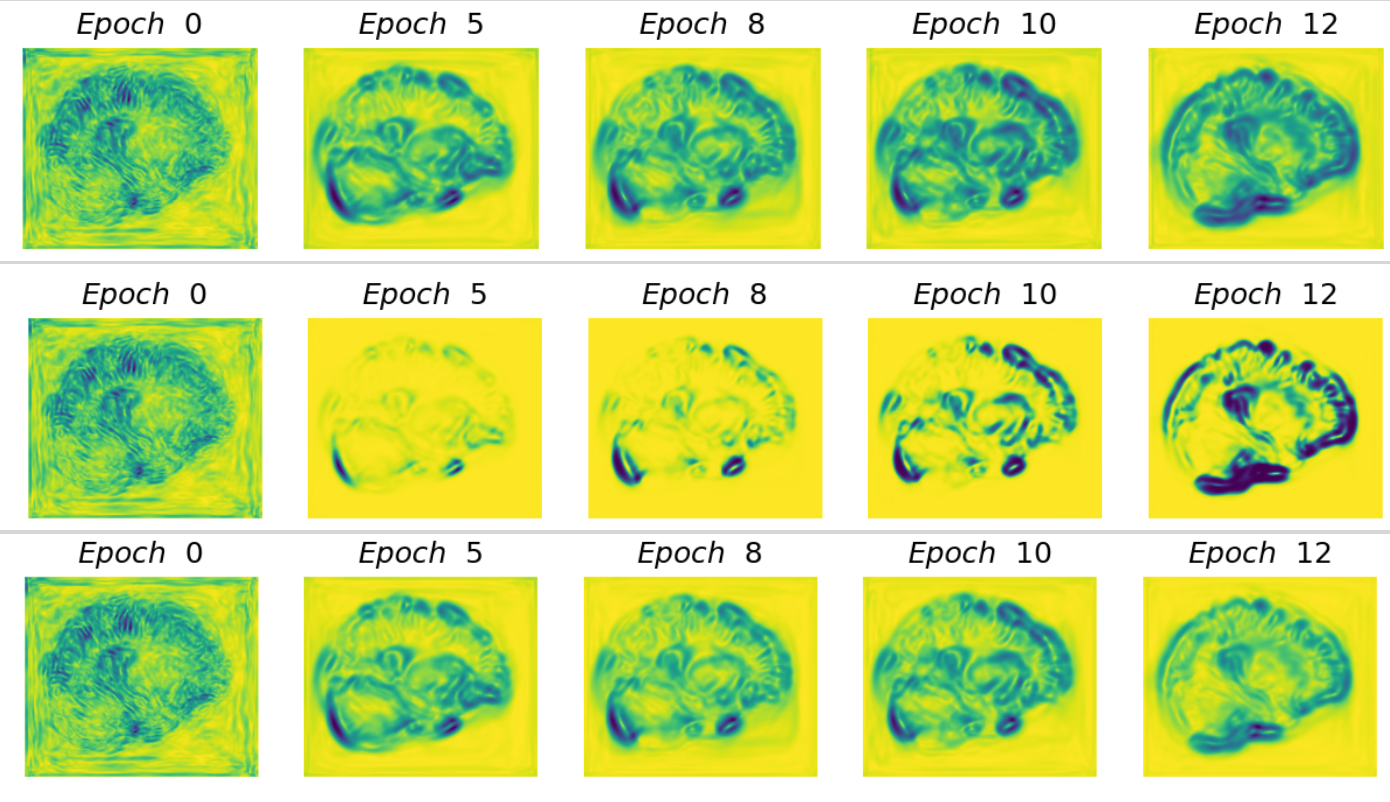}
\caption{Examples of dynamic adjustment of \(\lambda_{\textnormal{strain}}\) (top), \(\mu_{\textnormal{shear}}\) (middle), and \(\alpha\) (bottom) during training in response to the gradient norm.}
\label{fig:adjusted}
\end{figure}

The dynamic adjustment of   $\lambda_{\textnormal{strain}}$, $\mu_{\textnormal{shear}}$, $\alpha$ based on the gradient norm ensures that areas with high deformation gradients receive stronger regularization, promoting smoothness and stability. Figure \ref{fig:adjusted} illustrates the adaptive behavior. Conversely, areas with low gradients are less constrained, allowing for more flexibility in the deformation field.

\subsection{Proposed Model: DARE}

We now formulate DARE (Deformable Adaptive Regularization Estimator), a supervised registration framework based on a spatially adaptive extension of the elastic energy. It integrates the adaptive loss \eqref{eq:dare-reg} into the unsupervised registration functional \eqref{eq:unsupervised}. 
To prevent folding in the deformation field, we add an additional penalty term,
\(
\int_{\Omega} \left[ \max \big( 0, -\det(\nabla \deform) \big) \right]^2 \, dx,
\)
which quadratically penalizes negative Jacobian determinants.

Following the unsupervised learning setting, DARE constructs the displacement network \(\unet_\theta\) by minimizing  
\begin{multline} \label{eq:dare}
\run_{\textnormal{DARE}}(\theta) 
\triangleq
\sum_{i} \sun\big(\fixed_i, \moving_i \circ  \deform_{i,\theta} \big)
+ 
c \cdot \sum_{i} \int_{\Omega} \left[ \max \big( 0,  -\det(\nabla \deform_{i,\theta} ) \big) \right]^2 \, dx
\\
+ 
\sum_{i} \int_{\Omega} \left[ \wadapt(\disp_{i,\theta}) \cdot \left( \wvol(\disp_{i,\theta}) \cdot \trace(\strain_{i,\theta})^2 
+ \wshear(\disp_{i,\theta}) \cdot \norm{ \strain_{i,\theta} }_F^2 \right) \right] dx
\,,
\end{multline}
where we use the abbreviations
\[
\disp_{i,\theta} \triangleq \unet_\theta(\fixed_i, \moving_i), \quad 
\deform_{i,\theta} \triangleq \mathrm{Id} + \disp_{i,\theta}, \quad 
\strain_{i,\theta} \triangleq \frac{1}{2} \left( \nabla \disp_{i,\theta} + \nabla \disp_{i,\theta}^T \right)
\]
for the displacement field, the deformation field, and the associated strain tensor, respectively.  

The functions \(\wvol\), \(\wshear\), and \(\wadapt\) are defined in \eqref{eq:w-vol}, \eqref{eq:w-shear}, and \eqref{eq:w-adapt}, respectively, and depend on the parameters:
\begin{myitem}
\item \(\lambda_0\), \(\mu_0\): control the base elastic regularization,
\item \(\theta\): determines sensitivity to deformation gradients,
\item \(\beta\), \(\Delta\): control the gradient-based adjustments to regularization,
\item \(\tau\), \(\kappa\): define the shear modulus adaptation.
\end{myitem}
Together with the penalty weight \(c\), these form the hyperparameters of DARE.

The proposed DARE framework provides substantial advantages for deformable image registration by dynamically adjusting the regularization strength according to local deformation gradients. This adaptive mechanism allows greater flexibility in regions with low gradient magnitudes, enabling the deformation field to respond smoothly to subtle variations. In contrast, areas exhibiting high deformation gradients receive stronger regularization, thereby enhancing stability and preserving physical plausibility. The integration of gradient-based weighting and spatially adaptive parameters facilitates effective control over the regularization behavior, striking a balance between accuracy and smoothness across the deformation field. This design mitigates the risks of over-regularization in homogeneous regions and under-regularization in regions with sharp transitions, ultimately resulting in more robust and physiologically realistic deformation mappings.

\section{Experiment and Results}
\label{sec:results}

This section describes the experimental setup and presents the results of our study. We evaluated the proposed DARE framework using three datasets: atlas-to-patient, inter-patient, and a multimodal intra-patient brain MRI dataset. Each dataset is designed to address specific registration challenges. The proposed DARE method \eqref{eq:unsupervised} was compared against widely used regularization approaches, including total variation, diffusion, and bending energy regularizations, all implemented within the same unsupervised formulation by discretizing the respective components.  Additionally, we include comparisons with deep learning–based registration methods: VoxelMorph \cite{Balakrishnan2018}, CycleMorph \cite{Kim2021}, Swin-VoxelMorph \cite{zhu2022swin}, ViT-V-Net \cite{Chen2021}, TransMorph \cite{Chen2022}, XMorpher \cite{shi2022xmorpher}, Deformer \cite{chen2022deformer}, ModeT \cite{wang2023modet}, and TransMatch \cite{chen2023transmatch}.

To assess registration performance, we employed the following evaluation metrics: the Dice similarity coefficient (quantifying registration accuracy), the percentage of Jacobian determinants greater than one (\%$|J| \ge 1$, indicating tissue expansion), the percentage of Jacobian determinants less than zero (\%$|J| \le 0$, indicating anatomically invalid deformations), and strain energy (SE, measuring the smoothness of the deformation field).

\subsection{Datasets Description}

We used separate dataset for Atlas-to-Patient Registration, Inter-Patient Registration and Multimodal Intra-Patient Registration.

\begin{myitem}

\item \textbf{IXI (Atlas-to-Patient Registration):}
For atlas-to-patient registration, we utilized the IXI dataset. All scans represent healthy individuals, and we adopted the preprocessed dataset as described in \cite{Chen2022}. Our dataset was divided into randomly selected 30 scans for training, 20 for validation, and 115 for testing. The atlas-to-patient brain MRI used in this study was sourced from \cite{Kim2021}. Each volume was downsampled to a resolution of $160 \times 192 \times 224$ pixels, and registration accuracy was assessed using label maps of 29 anatomical structures.

\item \textbf{OASIS (Inter-Patient Registration):}
We employed the OASIS dataset, a publicly available resource provided through the 2021 Learn2Reg challenge \cite{Marcus2007, hoopes2021hypermorph}. The experiments utilized the preprocessed version of the dataset, as outlined in \cite{Chen2022}. Our workflow included 30 scans for training and 19 scans from the validation set for evaluation. The MRI scans, which have a resolution of $160 \times 192 \times 224$, were preprocessed to remove the skull, align volumes, and normalize intensities. Registration performance was evaluated using metrics such as the Dice Score, with 29 anatomical label masks serving as references.

\item \textbf{MUI-P (Multimodal Intra-Patient Registration):}
For Multimodal Intra-Patient Registration, we analyzed data provided by the Medical University of Innsbruck, Ethical approval number: UN5100 Sitzungsnummer:325/4.19. This dataset, referred to as MUI-P, contains 60 pairs of T1- and T2-weighted MRIs acquired in a whole body 3T MR-scanner (Siemens, Germany), using two different types of headcoils, one 64-channel 1H Hydrogen head-neck coil (Siemens, Erlangen, Germany) and one double-tuned 1H/31P Phosphorous volume head coil (Rapid Biomedical, Würzburg, Germany). From this set, 30 pairs were used for training, 10 for validation, and 20 for testing. These multimodal images allowed us to assess the registration method’s robustness across different modalities.

\end{myitem}

As a preprocessing step, all images are affinely aligned to ensure that any remaining misalignment between the volumes is due to nonlinear perturbations. This step isolates the nonlinear components of misalignment, enabling a more focused application of deformable registration techniques.

\subsection{Comparison with Variational Models}

We first compare DARE to a set of self-supervised variational models that share the same general structure as in Equation~\eqref{eq:unsupervised}, but differ in the form of the regularization term.

The regularizer \(\gun\) is chosen as either total variation (TV), diffusion, bending energy, or the proposed DARE method. TV penalizes the sum of gradient magnitudes to encourage piecewise smoothness, while bending energy penalizes second-order derivatives to enforce smooth curvature in the deformation field. Diffusion regularization encourages smooth spatial variations by penalizing first-order derivatives. For the similarity measure \(\sun\), we employ locally normalized cross-correlation, which is well suited for handling intensity variations commonly encountered in medical imaging. For multimodal MRI registration, we instead use local mutual information, a robust measure that captures the statistical dependence between images with differing intensity distributions. This setup enables reliable and accurate image alignment across diverse modalities.

For a fair comparison, we used the TransMorph architecture with configurations similar to those described in the original paper \cite{Chen2022}. Hyperparameter values used in this study were determined through grid search; Table~\ref{tab:hyperparameters} lists the selected values for DARE.

\begin{table}[htbp]
\centering
\caption{Hyperparameter values used in the experiments.}
\label{tab:hyperparameters}
\resizebox{0.6\textwidth}{!}{%
\begin{tabular}{|c|c|c|c|c|c|c|c|c|}
\toprule
$\lambda_0$ & $\mu_0$ & $c$ & $\beta$ & $\beta_0$ & center & scale & threshold \\
\midrule
1.0 & 0.5 & 10.0 & 1.0 & 1.0 & 0.05 & 0.01 & 0.1 \\
\bottomrule
\end{tabular}
}
\end{table}

\begin{myitem}
\item \textbf{Quantitative Comparison:}
Table \ref{tab:reg_methods_comparison} provides a detailed comparison of proposed DARE against state-of-the-art regularization techniques using Total Variation (TV), Diffusion, and Bending Energy.  DARE outperforms other regularization approaches for all datasets, achieving the highest Dice scores (0.821, 0.743, and 0.7963 for OASIS, IXI, and MUI-P, respectively) while maintaining low strain energy and reducing invalid deformations ($|J| \le 0$). Notably, DARE approach demonstrates significant improvements in the OASIS, IXI and MUI-P datasets, where it minimizes strain energy and ensures deformation validity (0.054, 0.001$\%$ and 0.011$\%$). These results highlight the effectiveness of DARE in achieving a better balance between registration accuracy and smoothness of the deformation field, validating its robustness across diverse datasets.

\begin{table}[htbp]
\caption{Quantitative Comparison across all datasets. Bold indicates best performance.}
\label{tab:reg_methods_comparison}
\centering
\resizebox{\textwidth}{!}{
\begin{tabular}{@{}lcccccccccccc@{}}
\toprule
\multirow{2}{*}{\textbf{Method}} 
& \multicolumn{4}{c}{\textbf{OASIS Dataset}} 
& \multicolumn{4}{c}{\textbf{IXI Dataset}} 
& \multicolumn{4}{c}{\textbf{MUI-P Dataset}} \\ 
\cmidrule(lr){2-5} \cmidrule(lr){6-9} \cmidrule(lr){10-13}
& Dice & \%$|J| \ge 1$ & \%$|J| \le 0$ & SE
& Dice & \%$|J| \ge 1$ & \%$|J| \le 0$ & SE
& Dice & \%$|J| \ge 1$ & \%$|J| \le 0$ & SE \\ 
\midrule
TV                & 0.794 & 52.481 & 1.454 & 0.182 & 0.629 & 40.762 & 2.973 & 0.294 & 0.743 & 46.476 & 0.521 & 0.054 \\
Diffusion         & 0.795 & 46.243 & 0.672 & 0.130 & 0.727 & 37.596 & 1.461 & 0.163 & 0.774 & 49.184 & 0.088 & 0.038 \\
Bending Energy    & 0.788 & 47.072 & 0.411 & 0.144 & 0.714 & 44.234 & 1.252 & 0.194 & 0.768 & 38.372 & 0.034 & 0.055 \\
DARE              & \textbf{0.821} & \textbf{48.673} & \textbf{0.002} & \textbf{0.058} 
                  & \textbf{0.743} & \textbf{49.081} & \textbf{0.001} & \textbf{0.042} 
                  & \textbf{0.796} & \textbf{61.081} & \textbf{0.001} & \textbf{0.011} \\ 
\bottomrule
\end{tabular}
}
\end{table}

\item \textbf{Qualitative Comparison:}  
Figures~\ref{fig:visual1} and~\ref{fig:visual2} present example results for Total Variation (TV), Diffusion, Bending Energy, and the proposed DARE regularization across all datasets. The results are shown as deformed volumes and corresponding displacement grids, with folding areas highlighted by red boxes. These red boxes emphasize regions prone to folding, where DARE produces smoother and more plausible deformations compared to the other methods. This demonstrates DARE’s effectiveness in minimizing folding artifacts and generating realistic deformation fields across diverse datasets.

\begin{figure}[htbp]
\centering

\begin{subfigure}{0.8\linewidth}
    \centering
    \includegraphics[width=\linewidth]{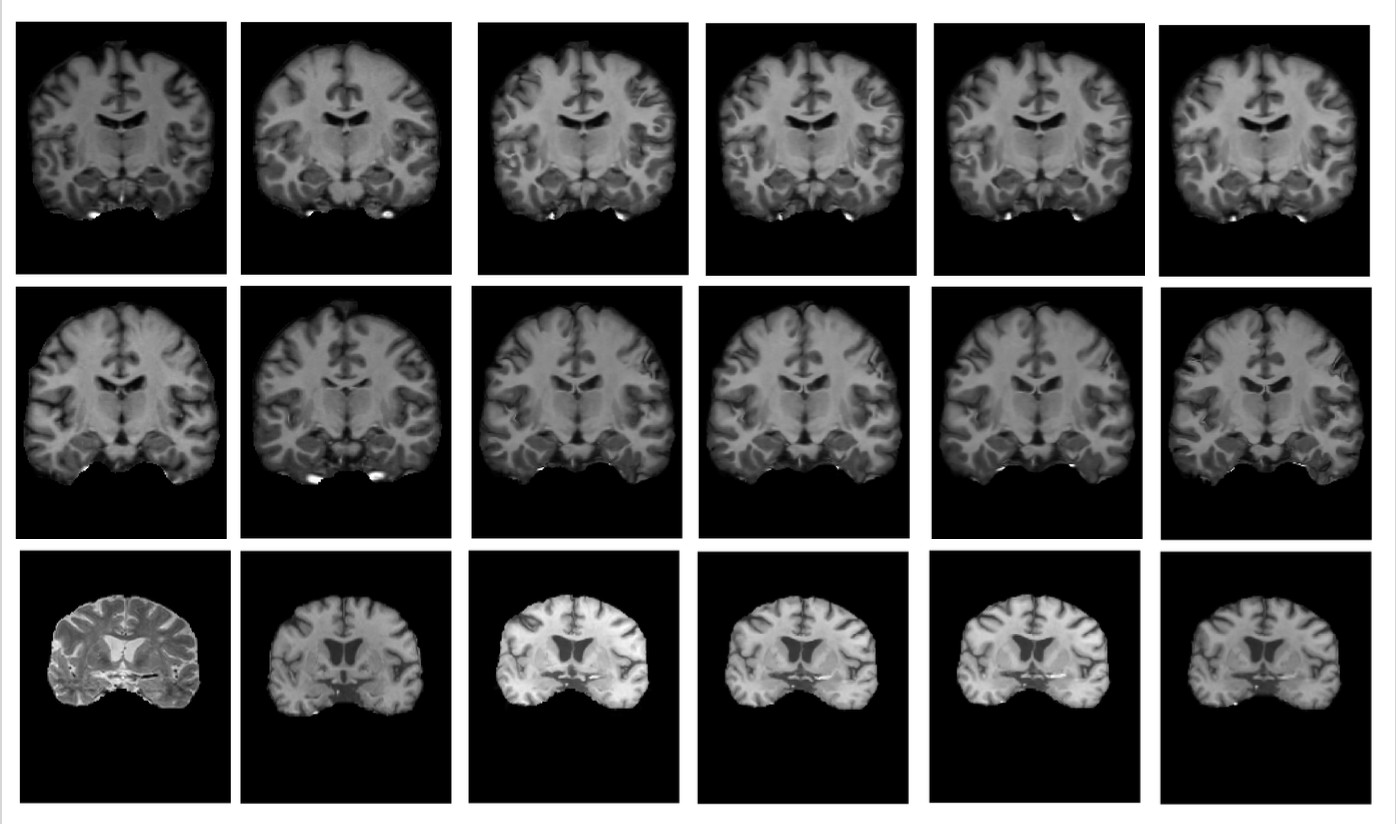}
    \caption{Deformed volumes from the IXI (top), OASIS (middle), and MUI-P (bottom) datasets. From left to right: fixed image, moving image, results with total variation, diffusion, bending energy regularizers, and DARE.}
    \label{fig:visual1}
\end{subfigure}

\vspace{0.5em}

\begin{subfigure}{0.8\linewidth}
    \centering
    \includegraphics[width=\linewidth]{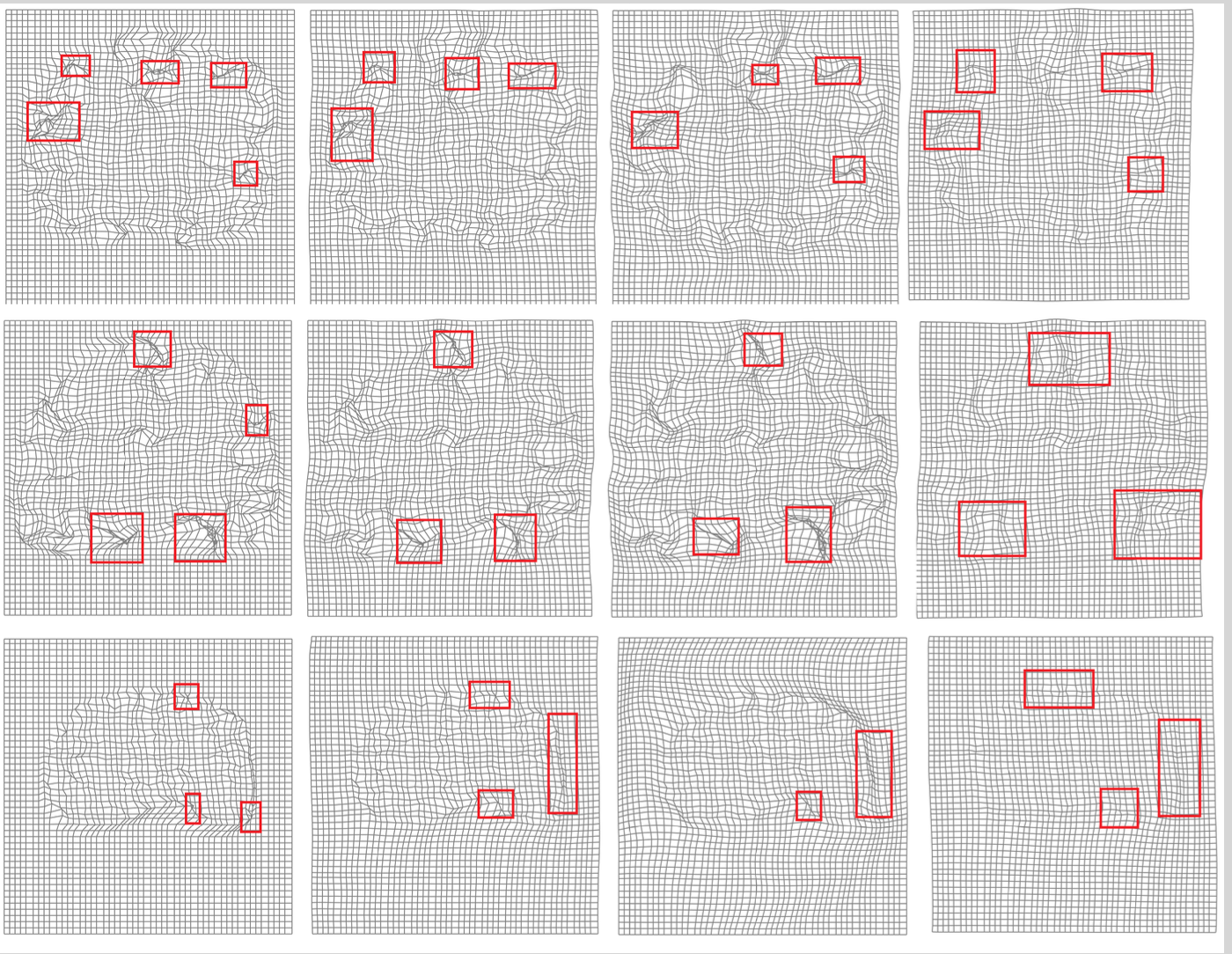}
    \caption{Displacement fields from the IXI (top), OASIS (middle), and MUI-P (bottom) datasets. From left to right: total variation, diffusion, bending energy regularizers, and DARE.}
    \label{fig:visual2}
\end{subfigure}

\caption{Qualitative comparison of (a) deformed volumes and (b) displacement fields.}
\label{fig:combined_visual}
\end{figure}

\item \textbf{Negative Jacobian determinants:}  
Figure~\ref{fig:jac} (top) shows the distribution of negative Jacobian determinants, providing insights into the extent and frequency of foldings present in the IXI dataset. The x-axis represents the magnitude of the foldings, while the y-axis indicates the number of affected voxels. Other methods exhibit a broader range of foldings, with Jacobian determinant values spanning from approximately -2.5 to -1.8. In contrast, DARE effectively restricts foldings to values above -0.4, with a significantly lower frequency of affected voxels. This demonstrates the superior ability of DARE to minimize foldings and maintain deformation regularity. Figure~\ref{fig:jac} (bottom) illustrates all negative Jacobian determinants across the IXI dataset, highlighting the locations of foldings in three-dimensional space. 

This visualization offers valuable insights into the spatial distribution of foldings within the dataset. Furthermore, it serves as a tool for comparing the effectiveness of the proposed method in preventing foldings, demonstrating its ability to minimize deformation irregularities.

\begin{figure}[htbp]
\centering
\includegraphics[width=0.8\linewidth]{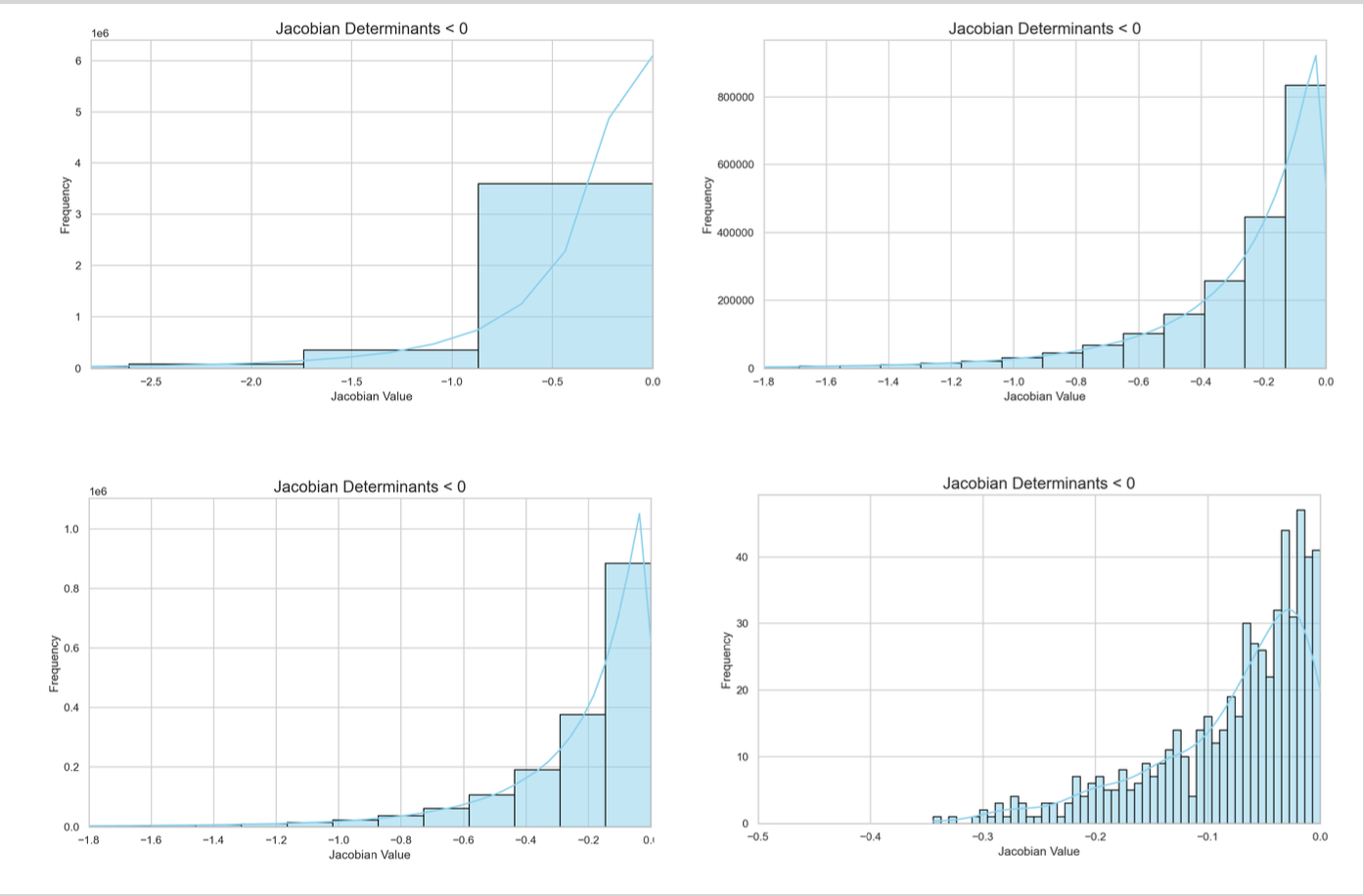}

\vspace{0.2em}

\includegraphics[width=0.8\linewidth]{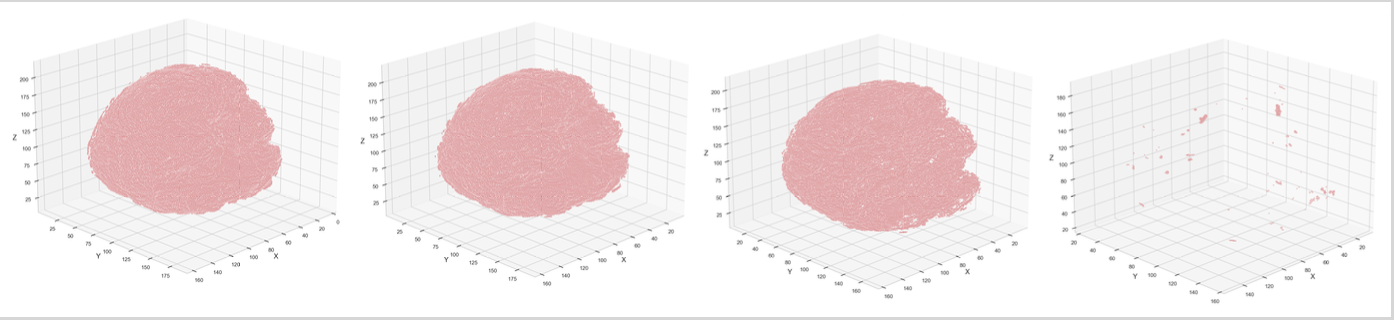}

\caption{Analysis of negative Jacobian determinants for the IXI dataset. Top: distribution of negative Jacobian determinants for different methods. Bottom: spatial locations of foldings visualized in 3D. From left to right: Total Variation, Diffusion, Bending Energy, and DARE.}
\label{fig:jac}
\end{figure}

\item \textbf{Analysis of strain:}  
Figure~\ref{fig:strain}(top) presents the distribution of strain across the dataset, highlighting the frequency and extent of strain values, including those exceeding 1, as produced by the different methods. Figure~\ref{fig:strain}(bottom) illustrates the strain applied to the IXI dataset by different methods. Strain, as a measure of deformation, indicates the extent to which tissue is stretched or compressed. In deformable image registration, strain values greater than 1 indicate significant deformation, where tissue has been excessively stretched. Such high strain can compromise the anatomical plausibility of the deformation and often correlates with negative Jacobian determinants, which signify undesirable folding or invalid deformations.  

High strain values may lead to inaccuracies in the registration process and potential loss of structural integrity in the deformation field. This visualization provides a comparative analysis of strain levels induced by popular methods, emphasizing regions and magnitudes of strain.  

\begin{figure}[htbp]
\centering
\includegraphics[width=0.8\linewidth]{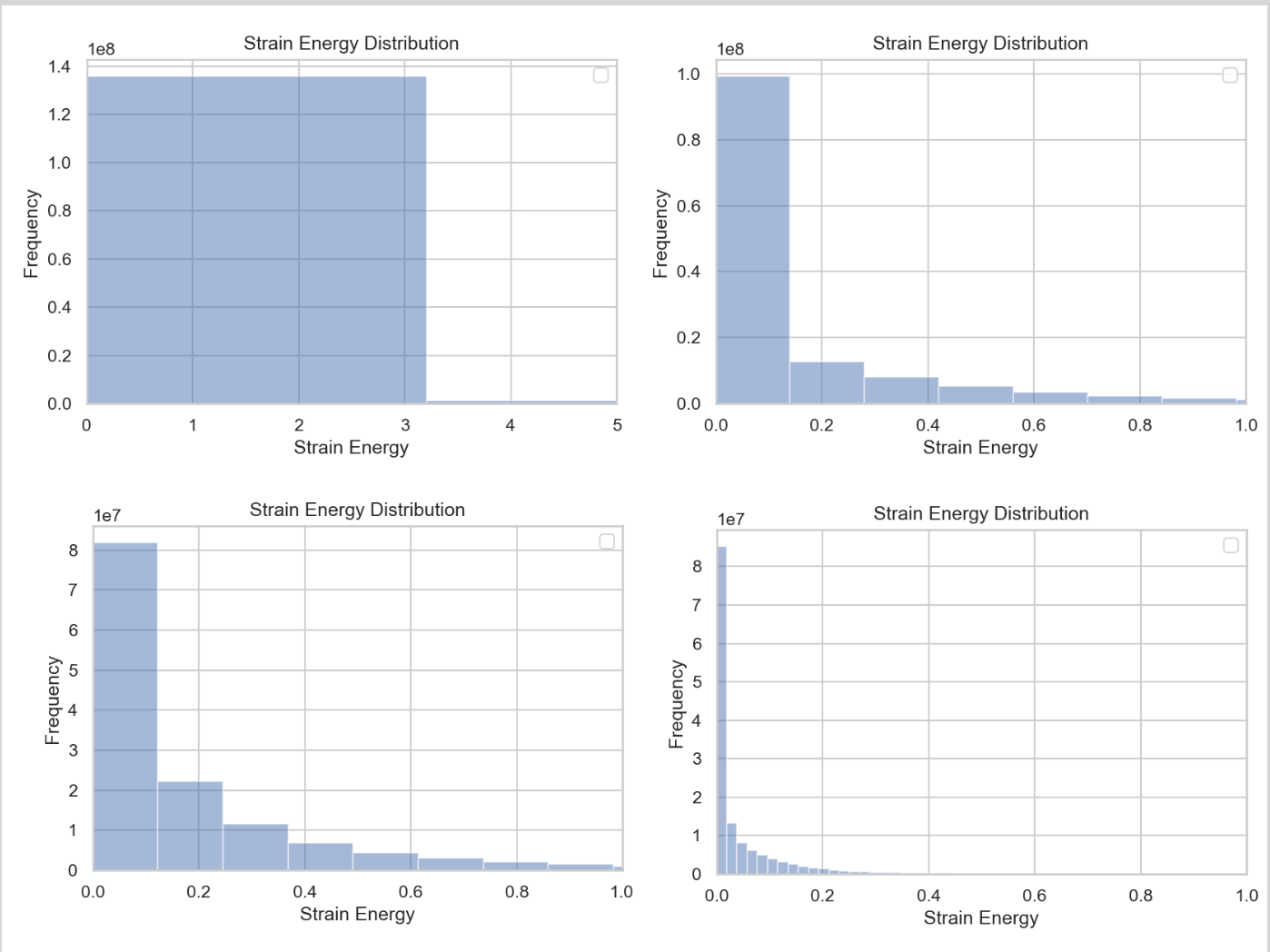}

\vspace{0.2em}

\includegraphics[width=0.8\linewidth]{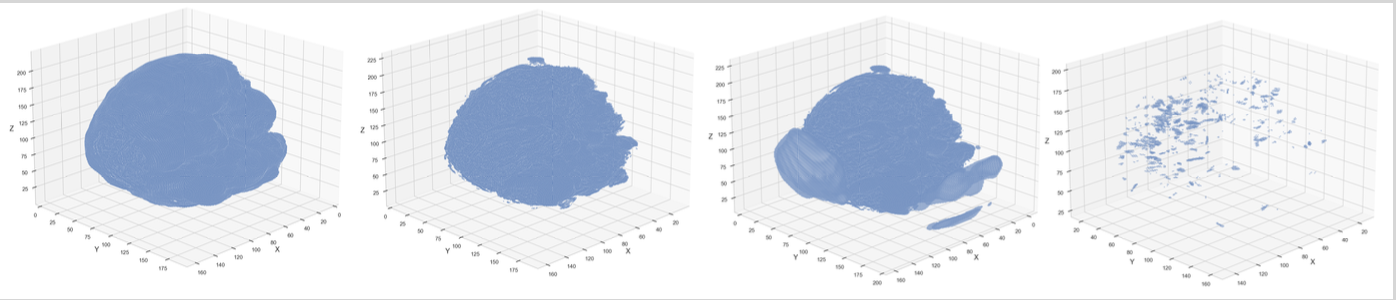}

\caption{Strain analysis across the IXI dataset. Top: distribution of strain values. Bottom: spatial visualization of strain regions where strain exceeds 1. From left to right: Total Variation, Diffusion, Bending Energy, and DARE.}
\label{fig:strain}
\end{figure}

\item \textbf{Analysis of volume changes:}  
Analyzing volume changes is crucial in registration tasks as it provides insights into the impact of deformation on anatomical structures. For this analysis, we selected five structures: Brain Stem, Thalamus, Amygdala, CSF, and Hippocampus. Table~\ref{tab:volcomparison} compares the percentage volume changes for different methods (TV, Diffusion, Bending Energy, DARE) on the OASIS and IXI datasets across these structures.  

DARE consistently achieves the lowest percentage volume change across all structures and datasets, indicating superior regularization performance. This highlights the effectiveness of the proposed method in minimizing deformation while preserving anatomical consistency. Figure~\ref{fig:volume_change} illustrates the percentage volume change for each structure. For clarity, 20 patients were randomly selected from the test dataset to represent the results in the figure. This analysis emphasizes the effect of registration on specific anatomical regions, providing valuable insight into the reliability and accuracy of the methods.

\begin{table}[htbp]
\caption{Comparison of volumetric changes (\%) across anatomical structures on the OASIS and IXI datasets.}
\label{tab:volcomparison}
\centering
\resizebox{0.8\textwidth}{!}{
\begin{tabular}{@{}lcccccccc@{}}
\toprule
Label Name & \multicolumn{4}{c}{\textbf{OASIS Dataset}} & \multicolumn{4}{c}{\textbf{IXI Dataset}} \\ 
\cmidrule(lr){2-5} \cmidrule(lr){6-9}
& TV & Diffusion & Bending Energy & DARE 
& TV & Diffusion & Bending Energy & DARE \\ 
\midrule
Thalamus    & 10.822 & 10.736 & 10.850 & 7.893  & 34.035 & 29.829 & 42.323 & 7.591 \\
Brain Stem  & 15.285 & 16.927 & 21.407 & 8.514  & 12.644 & 9.970  & 10.348 & 7.587 \\
Hippocampus & 10.081 & 9.830  & 10.281 & 6.731  & 17.541 & 15.325 & 16.464 & 6.588 \\
Amygdala    & 20.188 & 21.614 & 16.231 & 12.207 & 3.978  & 3.976  & 7.488  & 6.700 \\
CSF         & 17.393 & 17.056 & 17.317 & 11.937 & 14.818 & 14.144 & 13.962 & 10.414 \\
\bottomrule
\end{tabular}
}
\end{table}

\begin{figure}[htb]
\centering
\includegraphics[width=0.8\linewidth]{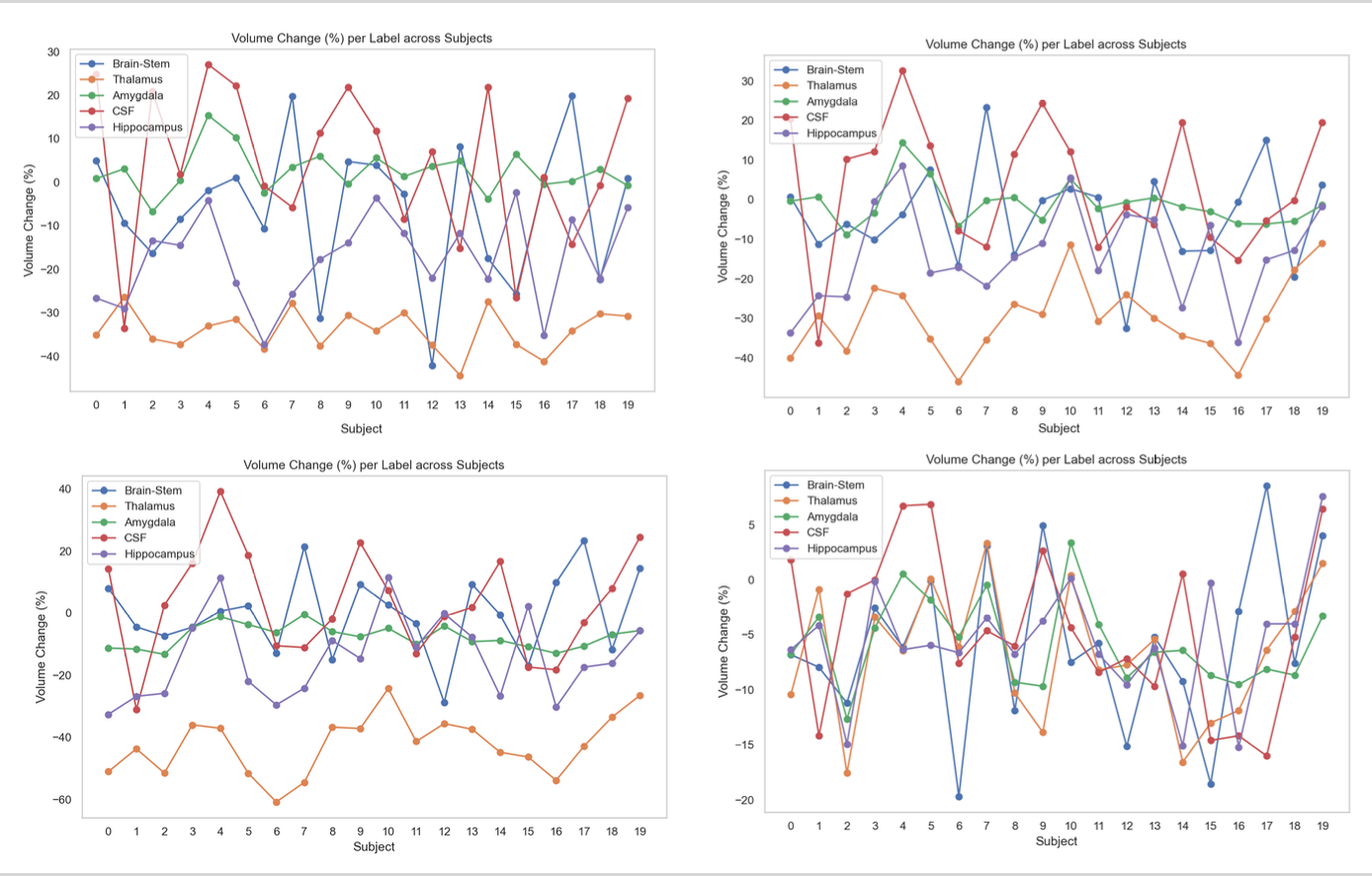}
\caption{Percentage of volume change in anatomical structures for 20 subjects of the IXI dataset using TV (top left), Diffusion (top right), Bending Energy (bottom left), and DARE (bottom right) regularizers.}
\label{fig:volume_change}
\end{figure}

\item \textbf{Evaluation of parameter adaptation}

Figure~\ref{fig:parameters1} illustrates the relationships between the adaptive parameters \(\lambda_{\textnormal{strain}}\) and \(\mu_{\textnormal{shear}}\) and the deformation gradient norm \(\norm{\nabla \disp}\), as well as their respective energy densities \(\energy_{\textnormal{strain}}\) and \(\energy_{\textnormal{shear}}\). In the top-left plot, \(\lambda_{\textnormal{strain}}\) decreases as \(\norm{\nabla \disp}\) increases, reflecting its exponential decay scaling, which reduces regularization in regions of high deformation gradients. Similarly, the top-right plot shows that \(\mu_{\textnormal{shear}}\) decreases with increasing \(\norm{\nabla \disp}\), exhibiting a smoother sigmoid-based adaptation. The bottom-left plot reveals an inverse correlation between \(\lambda_{\textnormal{strain}}\) and \(\energy_{\textnormal{strain}}\), where higher strain energy densities correspond to regions of reduced regularization weight. Finally, the bottom-right plot shows a similar inverse relationship between \(\mu_{\textnormal{shear}}\) and \(\energy_{\textnormal{shear}}\), allowing lower shear penalties in areas of significant shear deformation. Overall, these relationships highlight the adaptive nature of the loss function, which dynamically balances smoothness and deformation fidelity across the deformation field.

\begin{figure}[htb]
\centering
\includegraphics[width=0.7\linewidth]{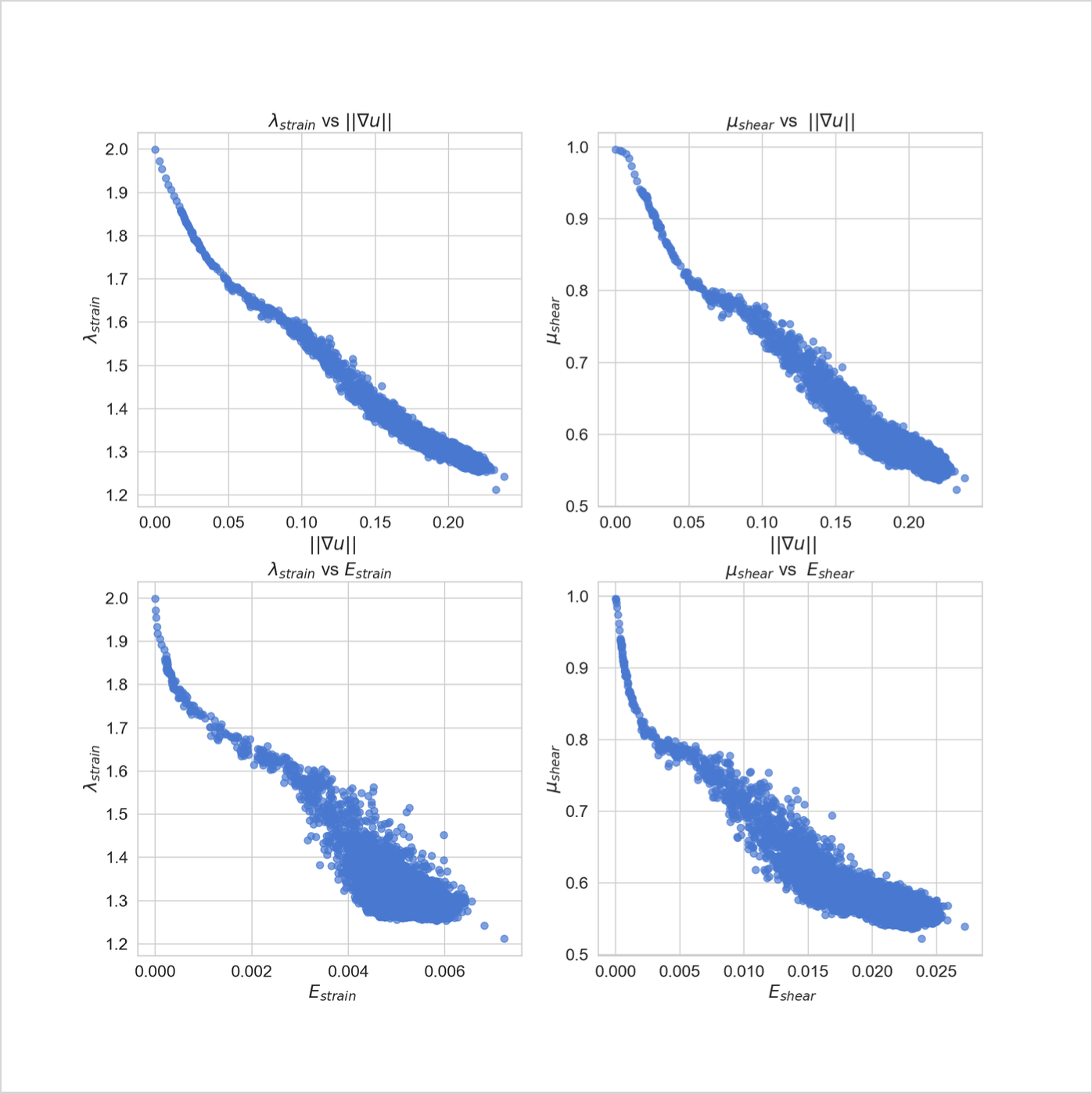}
\caption{Relationships between adaptive parameters (\(\lambda_{\textnormal{strain}}\) and \(\mu_{\textnormal{shear}}\)) and the deformation gradient norm \(\norm{\nabla \disp}\), as well as their respective energy densities \(\energy_{\textnormal{strain}}\) and \(\energy_{\textnormal{shear}}\).}
\label{fig:parameters1}
\end{figure}

Figure~\ref{fig:3d_lmE} illustrates the 3D relationship between \(\lambda_{\textnormal{strain}}\), \(\mu_{\textnormal{shear}}\), and the total energy density \(\energy_{\textnormal{total}} = \energy_{\textnormal{strain}}+\energy_{\textnormal{shear}}\). The scatter points indicate that as \(\energy_{\textnormal{total}}\) increases, both \(\lambda_{\textnormal{strain}}\) and \(\mu_{\textnormal{shear}}\) decrease, reflecting their adaptive behavior. This dynamic adjustment ensures lower regularization weights in regions of higher deformation energy, prioritizing flexibility where large deformations occur while maintaining smoothness in less deformed areas. The smooth decline along all axes indicates a consistent interplay between strain regularization, shear penalty, and total energy throughout the deformation field.

\begin{figure}[htbp]
\centering
\begin{tabular}{@{}l@{\hspace{1cm}}p{0.25\linewidth}@{}}
\includegraphics[width=0.5\linewidth]{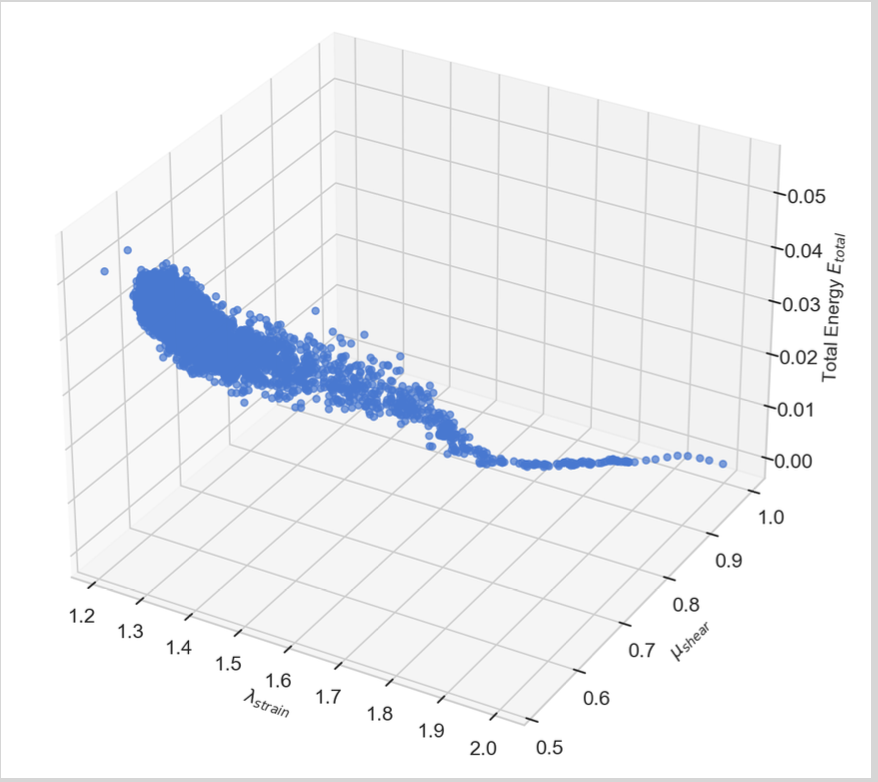} &
  \parbox[b]{\linewidth}{
    \caption{3D relationship between \(\lambda_{\textnormal{strain}}\), \(\mu_{\textnormal{shear}}\), and total energy \(\energy_{\textnormal{total}}\), highlighting adaptive regularization behavior.}
    \label{fig:3d_lmE}
  }
\end{tabular}
\end{figure}

Figure~\ref{fig:parameters2} presents four scatter plots describing relationships among adaptive parameters (\(\lambda_{\textnormal{strain}}\), \(\mu_{\textnormal{shear}}\)), folding penalty, and energy components during deformable image registration. The top-left plot shows an inverse relationship between \(\lambda_{\textnormal{strain}}\) and \(\energy_{\textnormal{total}}\), where \(\lambda_{\textnormal{strain}}\) decreases as total energy increases, consistent with its dynamic adjustment based on deformation gradients. Similarly, the top-right plot illustrates \(\mu_{\textnormal{shear}}\) versus \(\energy_{\textnormal{total}}\), showing a comparable decreasing trend due to sigmoid-based adaptation. The bottom-left plot depicts folding penalty versus \(\energy_{\textnormal{total}}\), indicating a rapid increase in folding penalties with rising total energy, suggesting higher risks of folding in regions with larger deformations. Finally, the bottom-right plot reveals a nonlinear relationship between \(\energy_{\textnormal{strain}}\) and \(\energy_{\textnormal{shear}}\), where trace energy oscillates as shear energy increases, reflecting the interplay between volumetric and shear deformations. Together, these plots highlight the adaptive loss function’s ability to balance deformation energy while penalizing topological inconsistencies.

\begin{figure}[htbp]
\centering
\includegraphics[width=0.8\linewidth]{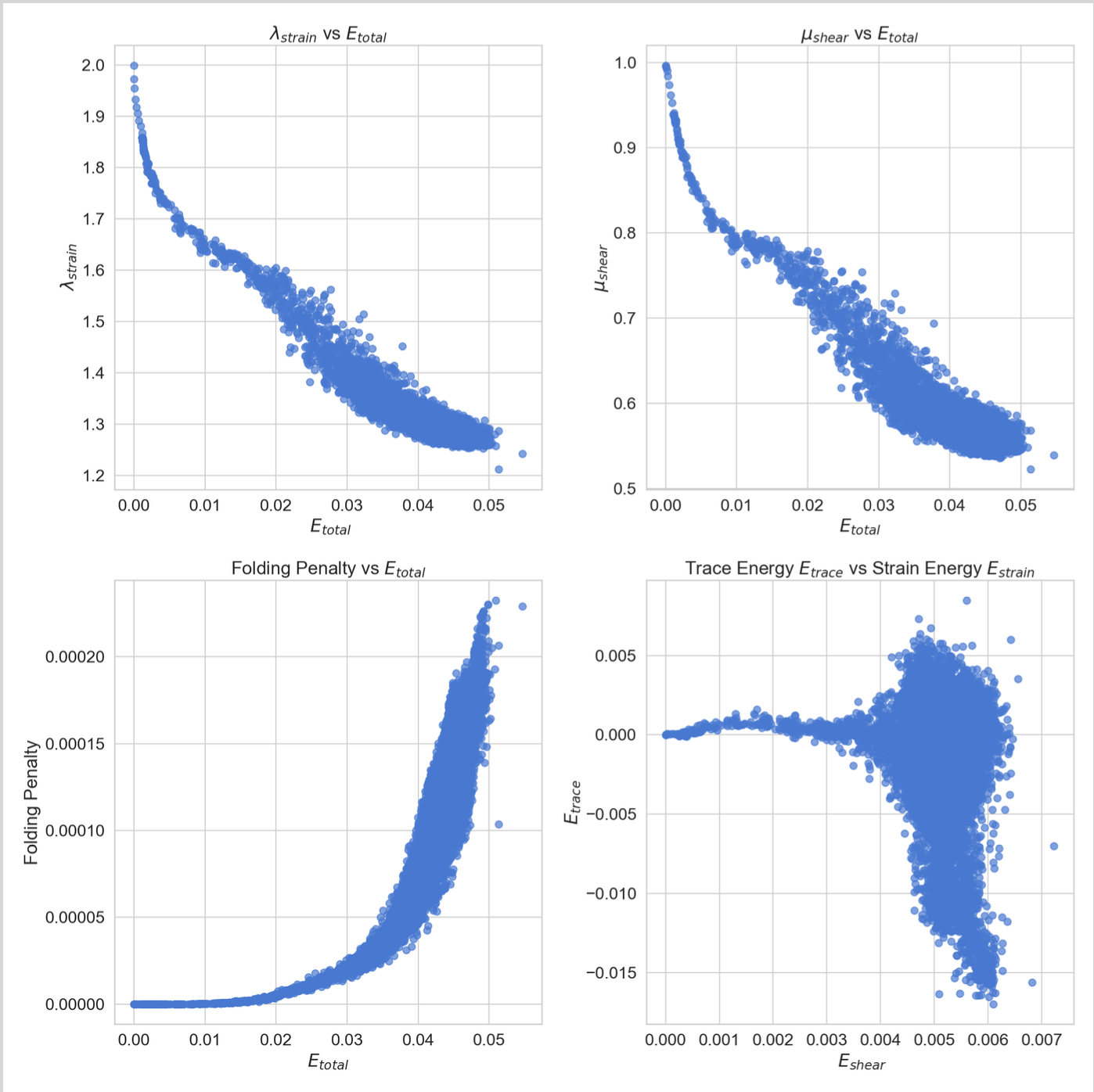}
\caption{Relationships between adaptive parameters, folding penalty, and energy components, highlighting dynamic regularization and deformation behavior.}
\label{fig:parameters2}
\end{figure}

\end{myitem}

\subsection{Comparison with Deep Learning Models}

Table~\ref{table_results} presents a comprehensive comparison of DARE against recent learned registration methods on the IXI, OASIS, and MUI-P datasets. Among all evaluated methods, the proposed approach consistently outperforms alternatives, achieving the highest Dice scores—0.743, 0.827, and 0.796 for the IXI, OASIS, and MUI-P datasets, respectively—alongside the lowest percentages of non-positive Jacobian determinants ($|J| \le 0$), registering values below 0.001\%, 0.002\%, and 0.001\%, respectively. These results underscore DARE's superior registration accuracy and deformation regularization.

\begin{table}[htbp]
\caption{Comparison of Registration Methods Across IXI, OASIS, and MUI-P Datasets}
\label{table_results}
\centering
\resizebox{\textwidth}{!}{%
\begin{tabular}{@{}lcccccc@{}}
\toprule
\textbf{Methods} 
& \multicolumn{2}{c}{\textbf{IXI Dataset}} 
& \multicolumn{2}{c}{\textbf{OASIS Dataset}} 
& \multicolumn{2}{c}{\textbf{MUI-P Dataset}} \\ 
\cmidrule(lr){2-3} \cmidrule(lr){4-5} \cmidrule(lr){6-7}
& Dice & \%$|J| \le 0$ 
& Dice & \%$|J| \le 0$ 
& Dice & \%$|J| \le 0$ \\ 
\midrule
VoxelMorph~\cite{Balakrishnan2018} & 0.701 (0.026) & 0.911 (0.142) & 0.751 (0.021) & 0.861 (0.132) & 0.742 (0.022) & 0.621 (0.192) \\
CycleMorph~\cite{Kim2021} & 0.709 (0.023) & 0.882 (0.164) & 0.762 (0.026) & 0.813 (0.261) & 0.751 (0.026) & 0.661 (0.211) \\
Swin-VoxelMorph~\cite{zhu2022swin} & 0.714 (0.027) & 0.822 (0.261) & 0.778 (0.028) & 0.998 (0.231) & 0.763 (0.022) & 0.414 (0.162) \\
ViT-V-Net~\cite{Chen2021} & 0.681 (0.031) & 0.961 (0.241) & 0.766 (0.024) & 0.786 (0.246) & 0.752 (0.029) & 0.792 (0.246) \\
TransMorph~\cite{Chen2022} & 0.723 (0.012) & 1.466 (0.223) & 0.783 (0.022) & 0.672 (0.222) & 0.774 (0.011) & 0.088 (0.021) \\
XMorpher~\cite{shi2022xmorpher} & 0.726 (0.021) & 0.868 (0.182) & 0.779 (0.024) & 0.895 (0.243) & 0.776 (0.024) & 0.124 (0.061) \\
Deformer~\cite{chen2022deformer} & 0.724 (0.026) & 0.772 (0.161) & 0.783 (0.014) & 0.672 (0.168) & 0.761 (0.022) & 0.321 (0.091) \\
ModeT~\cite{wang2023modet} & 0.729 (0.026) & 0.034 (0.013) & 0.775 (0.012) & 0.089 (0.012) & 0.779 (0.017) & 0.042 (0.012) \\
TransMatch~\cite{chen2023transmatch} & 0.728 (0.013) & 0.791 (0.262) & 0.791 (0.022) & 0.721 (0.233) & 0.764 (0.026) & 0.012 (0.008) \\
\textbf{DARE (Ours)} & \textbf{0.743 (0.016)} & \textbf{$<$ 0.001 ($<$ 0.001)} 
& \textbf{0.827 (0.018)} & \textbf{0.002 (0.001)} 
& \textbf{0.796 (0.011)} & \textbf{0.001 ($<$ 0.001)} \\ 
\bottomrule
\end{tabular}%
}
\end{table}

Other notable performers include ModeT and TransMatch, which achieved competitive Dice scores but exhibited significantly higher percentages of $|J| \le 0$ compared to DARE, indicating less robust control over deformation regularity. Traditional CNN-based methods such as VoxelMorph and CycleMorph yielded moderate Dice scores but suffered from high incidences of non-physical deformations, particularly on the IXI and MUI-P datasets. Similarly, methods like Swin-VoxelMorph and XMorpher delivered relatively strong Dice metrics but were compromised by elevated rates of $|J| \le 0$, suggesting challenges in maintaining anatomical plausibility.

In summary, the proposed method demonstrates the most balanced performance among all compared approaches—achieving high registration accuracy while preserving anatomical consistency through effective regularization. This makes DARE a robust and reliable choice for medical image registration, with strong generalizability across datasets with varying anatomical characteristics.

\section*{Ethics Approval Statement}

This study includes data provided by the Medical University of Innsbruck. Ethical approval for the use of this data was granted by the institutional review board under approval number: UN5100, Sitzungsnummer: 325/4.19.

\section*{Acknowledgment}

This work was supported by the Austrian Science Fund (FWF) [grant number DOC 110]
\section{Conclusion}
\label{sec:conclusion}

This study addresses the limitations of traditional regularization strategies in learned deformable image registration by introducing a context-aware approach that adaptively modulates regularization parameters. By dynamically adjusting the elastic regularization weights based on the gradient norm of the deformation field, the proposed method enables spatially and contextually responsive regularization. This mechanism balances deformation flexibility and stability by incorporating strain and shear energy components that adapt to local variations—an essential feature for high-precision medical imaging tasks. Another key component of the method is a folding prevention mechanism that penalizes negative Jacobian determinants, ensuring smooth, physically plausible deformations and reducing the risk of non-physical artifacts such as folding. This prevents over-smoothing in regions that require fine alignment, while preserving anatomical consistency in more homogeneous areas, thereby maintaining critical details like small lesions and fine structures.

By overcoming the rigidity of uniform regularization, the proposed approach enhances the accuracy, robustness, and anatomical realism of learning-based registration models. Its adaptability and sensitivity to local context make it particularly well-suited for clinical applications where both precision and consistency are essential. Overall, this work demonstrates the potential of dynamically adaptive regularization frameworks to improve the reliability and effectiveness of medical image registration.

\end{document}